\documentclass[sigconf]{acmart}

\usepackage{microtype}
\usepackage{graphicx}

\AtBeginDocument{%
  }

\makeatletter
\let\acmart@orig@mktitle@iii\@mktitle@iii
\def\@mktitle@iii{\hsize=\textwidth
    \setbox\mktitle@bx=\vbox{%
      \noindent\makebox[\hsize]{%
        \includegraphics[height=0.95cm]{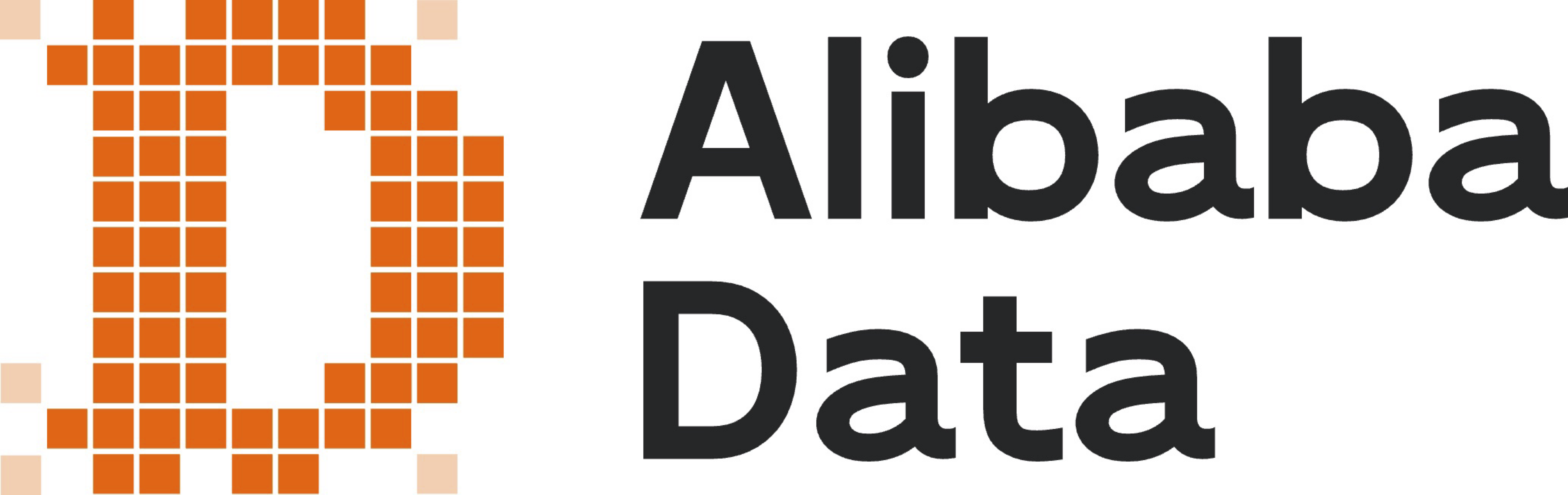}%
        \hfill
        \includegraphics[height=1.45cm]{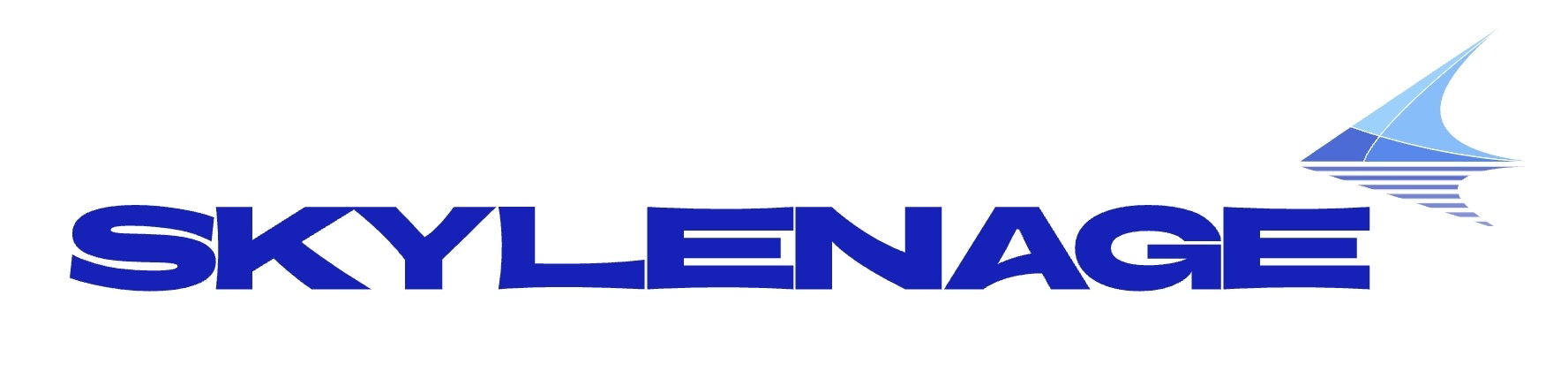}%
      }\\[-0.2em]
      \noindent\rule{\hsize}{0.5pt}\\[0.4em]
      \@titlefont\centering
      \@ACM@title@width=\hsize
      \parbox[t]{\@ACM@title@width}{\centering\@titlefont
        \@title\@translatedtitle%
        \ifx\@subtitle\@empty\else
          \par\noindent{\@subtitlefont\@subtitle\@translatedsubtitle}
        \fi
      }%
      \par\bigskip}}%
\makeatother

\setcopyright{acmlicensed}
\copyrightyear{2026}
\acmYear{2026}
\acmDOI{XXXXXXX.XXXXXXX}
\acmConference[KDD]{KDD 2026 Datasets and Benchmarks Track }{Aug 06 2026}{Jeju, Korea}
\acmISBN{978-1-4503-XXXX-X/2018/06}

\begin{document}


\title{FeynmanBench: Benchmarking Multimodal LLMs on Diagrammatic Physics Reasoning}


\author{Zeyu Wang}
\authornote{Both authors contributed equally to this research.}
\affiliation{%
  \institution{Alibaba Group}
  \city{Beijing}
  \country{China}
}
\email{chenfan.wzy@alibaba-inc.com}

\author{Jingye Xu}
\authornotemark[1]
\affiliation{%
	\institution{Alibaba Group}
	\city{Beijing}
	\country{China}}
\email{xjy02516377@alibaba-inc.com}

\author{Xiaogang Li}
\affiliation{%
	\institution{Alibaba Group}
	\city{Beijing}
	\country{China}}
\email{lixiaogang.lxg@alibaba-inc.com}

\author{Peiyao Xiao}
\affiliation{%
	\institution{Alibaba Group}
	\city{Beijing}
	\country{China}}
\email{xiaopeiyao.xpy@alibaba-inc.com}

\author{Qinhao Kong}
\affiliation{%
	\institution{Skylenage}
	\city{Beijing}
	\country{China}
}
\email{kongqinhao.kqh@alibaba-inc.com}

\author{Ben Wang}
\affiliation{%
  \institution{Alibaba Group}
  \city{Beijing}
  \country{China}
}
\email{yuanjian.wb@alibaba-inc.com}

\author{Chengliang Xu}
\affiliation{%
  \institution{Alibaba Group }
  \city{Beijing}
  \country{China}}
\email{xiaodu.xcl@alibaba-inc.com}

\author{Zichao Chen}
\affiliation{%
	\institution{Alibaba Group}
	\city{Beijing}
	\country{China}}
\email{chenzichao.czc@alibaba-inc.com}

\author{Bing Zhao}
\authornote{Corresponding Author.}
\affiliation{%
	\institution{Alibaba Group}
	\city{Beijing}
	\country{China}}
\email{xiongdao@alibaba-inc.com}

\author{Hu Wei }
\authornotemark[2]
\affiliation{%
  \institution{Alibaba Group}
  \city{Beijing}
  \country{China}}
\email{kongwang@alibaba-inc.com}

\renewcommand{\shortauthors}{Wang et al.}

\begin{abstract}
Current multimodal benchmarks for scientific reasoning primarily evaluate local information extraction---models recognize symbols and values and then perform textual inference. They do not assess whether models can reason over the global structural properties of formal diagrams, such as topology, conservation constraints, and the consistent mapping between visual patterns and algebraic expressions. We introduce FeynmanBench, a benchmark of over 2,000 tasks centered on Feynman diagrams spanning the electromagnetic, weak, and strong interactions of the Standard Model. Each instance couples a diagram image with minimal textual conventions and requires models to recover the full physical content---vertex inventory, propagator types, topological connectivity, momentum routing, and the complete scattering amplitude. An automated generation and verification pipeline produces the diagrams, annotations, and reference answers under standardized rules. Evaluating 19 state-of-the-art multimodal LLMs, we find a consistent failure pattern: models achieve 70--95\% on local recognition (vertex and propagator identification) but collapse to 13--17\% on topological reconstruction (CP3), and near zero on full algebraic derivation (CP5). FeynmanBench offers a controlled testbed for multimodal reasoning over formal scientific diagrams and highlights fundamental limitations of current architectures in topology-sensitive scientific reasoning.
\end{abstract}


\maketitle

\section{Introduction}

Diagrammatic reasoning is a core methodology in modern physics and mathematics~\cite{olivier2001diagrammatic}. Among diagrammatic formalisms, Feynman diagrams~\cite{feynman1949spacetime,kumericki2016feynman} play a unique role: they encode perturbative expansions in quantum field theory as graph structures, mapping intuitive visual representations to rigorous computation via unambiguous Feynman rules. Each diagram is a graph whose connectivity, conservation laws, and algebraic derivation must satisfy precise global constraints. Feynman diagrams are especially suitable for benchmarking because they combine three properties: a rigid visual syntax (vertices, propagators, and external legs follow fixed graphical conventions), global physical constraints (conservation laws, gauge symmetries, and the topological identity $L = I - V + 1$), and machine-verifiable symbolic outputs (amplitudes can be generated and checked automatically). Unlike many scientific images that admit descriptive interpretation, a Feynman diagram has tightly constrained semantics where small topological differences produce different physical processes and amplitudes. This makes them an ideal testbed for evaluating whether multimodal models can perform structured visual reasoning rather than superficial pattern matching.

Existing multimodal benchmarks in the physical sciences~\cite{he-etal-2024-olympiadbench,wang2025physunibench,feng2025physics,chen2023theoremqa} predominantly follow a ``recognition then reasoning'' paradigm: models first extract discrete symbols via OCR or object recognition, then perform textual derivation using internalized formulas. While useful for measuring parameter extraction and textbook problem-solving, this paradigm does not test whether models can reason directly over the global structure of a diagram. Topological relationships, cross-regional constraints, and implicit physical symmetries---the defining features of formal scientific notation---remain unexamined~\cite{xu2025visulogic,altahan2024unibench}. Recent benchmarks such as PhyX~\cite{shen2025phyx} and MAPS~\cite{zhu2025maps} reduce textual dependence or introduce physics engines, but still either generate intermediate descriptions or shift evaluation toward tool-calling accuracy.

We introduce FeynmanBench, a benchmark designed to evaluate multimodal LLMs on tasks that require simultaneous diagrammatic recognition, topological reasoning, and formal algebraic derivation directly in the visual domain. The benchmark comprises over 2,000 Feynman diagrams drawn from more than 100 distinct topological types, covering electromagnetic, electroweak, and strong interactions. Each instance provides a diagram image together with minimal textual conventions (interaction type, external leg identities, and momentum orientation)---sufficient only to resolve drawing-convention ambiguities, not to supply topological or algebraic answers. Models must report the vertex inventory (CP1), propagator content (CP2), topological connectivity (CP3), momentum routing (CP4), and the full scattering amplitude with symmetry factors (CP5). An automated pipeline based on FeynArts and FeynCalc~\cite{hahn2001feynarts3,shtabovenko2016feyncalc90} generates the diagrams, topological annotations, and reference amplitudes under standardized gauge conventions, ensuring scalability and verifiability.

In summary, our core contributions are: (1)~FeynmanBench, the first benchmark for evaluating multimodal diagrammatic reasoning over Feynman diagrams; (2)~an automated generation and verification pipeline that produces diagrams, annotations, and reference amplitudes under standardized conventions; and (3)~a systematic evaluation of 19 MLLMs exposing the gap between local visual recognition and global structural reasoning. Our experiments reveal a consistent bottleneck: local symbol recognition does not translate into global structural understanding. The top-scoring model achieves 52.5\% overall, while the strongest full-benchmark models plateau around 28--30\%. Across all models, we observe a consistent collapse at CP3 (topological connectivity)---even models with 94.7\% vertex accuracy drop to 16.8\% on topology---and CP5 (algebraic/symmetry factors) remains below 2\% for most. Strikingly, we find that reducible topologies are consistently harder than irreducible ones, suggesting that models rely on memorized textbook patterns rather than transferable structural deduction. Our code, pipeline, and benchmark will be released upon acceptance.

\section{Related Work}

\textbf{Multimodal Scientific Benchmarks.}
Benchmarks for multimodal scientific reasoning have progressed from undergraduate textbooks to Olympiad-level problems~\cite{he-etal-2024-olympiadbench,feng2025physics}. TheoremQA~\cite{chen2023theoremqa} and PhysUniBench~\cite{wang2025physunibench} require models to extract numerical parameters from images and apply formulas, while MM-PhyQA~\cite{anand2024mmphyqa} uses multi-image chain-of-thought prompting. A consistent limitation unites these efforts: evaluation focuses on the combined accuracy of parameter extraction and textual logic, leaving the visual reasoning component unmeasured. PhyX~\cite{shen2025phyx} reduces textual cues to isolate visual dependence, but still allows models to generate structured descriptions as an intermediate step. MAPS~\cite{zhu2025maps} and LLMPhy~\cite{cherian2024llmphy} integrate physical simulators, shifting evaluation toward tool use rather than intrinsic visual reasoning. In parallel, recent visual multimodal models have increasingly unified understanding and generation within single architectures~\cite{hu2025unified_mmugm,deng2025_bagel_emerging_properties,wu2025_vila_u}, yet systematic evaluation of visual reasoning over formal scientific notation remains under-explored. In contrast, FeynmanBench requires models to reason directly over diagrammatic structure, where correctness is determined by global topology and algebraic consistency rather than local feature extraction. We designed the benchmark so that even a model with perfect OCR capability cannot succeed without genuine visual-structural understanding.

\textbf{ML for Feynman Diagrams.}
Prior work applies machine learning to accelerate Feynman-diagram computation---predicting scattering matrix elements via GNNs~\cite{Mitchell2022LearningFeynmanGNN}, solving Feynman integrals with PINNs~\cite{Calisto2024LearningFeynmanDE}, and sampling diagrams via normalizing flows~\cite{Leoni2024GlobalSamplingNF}. These efforts use ML as a computational tool for physics, but do not benchmark multimodal reasoning over the diagrams themselves. Our work is orthogonal: rather than using ML to solve physics problems, we use physics problems as a probe for ML reasoning capability.

\section{FeynmanBench}

\subsection{Task Definition}

Each instance in FeynmanBench consists of two inputs: (1)~a Feynman diagram image, and (2)~minimal textual conventions specifying the interaction type (e.g.,~QED, weak, QCD), the identities and momentum labels of external legs, and the diagram orientation. These conventions are intentionally limited---they resolve drawing ambiguity (e.g.,~momentum-direction conventions for antiparticles) without revealing any topological or algebraic answer. The model must produce five outputs, evaluated via checkpoints CP1--CP5:

\noindent\textbf{CP1 -- Vertex inventory:} count and type of each interaction vertex (e.g.,~$\bar{\psi}\gamma^\mu\psi A_\mu$, $W^+W^-Z$), with the correct linked field count per vertex.

\noindent\textbf{CP2 -- Propagator inventory:} complete list of internal lines with correct particle types and multiplicities.

\noindent\textbf{CP3 -- Topological connectivity:} graph of vertex connections isomorphic to the original diagram.

\noindent\textbf{CP4 -- Momentum routing:} self-consistent internal momentum assignment satisfying conservation at each vertex.

\noindent\textbf{CP5 -- Full amplitude:} algebraic expression including Dirac structures, trace contractions, Lorentz index closures, global signs, fermion-loop factors, and combinatorial symmetry factors.

\noindent The five checkpoints form a deliberate progression. CP1 and CP2 test whether the model can identify primitives; CP3 and CP4 test whether it can assemble those primitives into a self-consistent physical graph; CP5 tests whether that graph-level understanding is sufficient to support full symbolic derivation. Each checkpoint is evaluated independently and unconditionally (a model need not pass earlier checkpoints to be scored on later ones), but in practice errors cascade: an incorrect vertex in CP1 invalidates the topology in CP3, which in turn makes the amplitude in CP5 unsalvageable. The overall model score is the arithmetic mean of the five unconditional checkpoint pass rates.

These conventions are necessary because certain drawing choices in professional Feynman diagrams are not semantically unique in isolation---momentum-arrow conventions for antiparticles, omitted charge labels for some bosons, and similar notational ambiguities arise in multi-purpose diagrams. The text therefore disambiguates notation without supplying structural or algebraic answers: the critical reasoning for CP3--CP5 remains entirely visual-structural.

\subsection{Dataset Generation}

We construct FeynmanBench using an automated pipeline built on FeynArts and FeynCalc~\cite{hahn2001feynarts3,shtabovenko2016feyncalc90}. We first enumerate connected graph topologies (vertex-line skeletons) with specified numbers of external legs and loops. We then insert physical fields by mapping particle content from Standard Model definition files onto the skeleton lines, enforcing vertex-level conservation laws and coupling rules. Finally, we convert field-inserted diagrams to analytic amplitudes by applying symbolic Feynman rules---propagators for internal lines, vertex factors for interaction points, and combinatorial symmetry factors from topological automorphisms. All amplitudes are generated in Feynman gauge.

For each diagram, we store: (1)~the rendered image; (2)~external-leg metadata; (3)~vertex-level type and connectivity annotations; (4)~internal propagator annotations; (5)~a canonicalized topology record; and (6)~the reference amplitude expression. This structured representation supports automated evaluation across all checkpoints. From the full candidate pool of over 200,000 diagrams, we select the benchmark subset via stratified sampling over topology class (A1--A5) and interaction sector (B1--B6), removing near-duplicate layouts to maximize structural diversity under a fixed evaluation budget.

\subsection{Data Taxonomy}

Diagrams are classified along two dimensions:
\begin{itemize}
    \item \textbf{Topology complexity (A1--A5):} A1: tree-level; A2: one-loop one-particle-reducible (1PR); A3: one-loop one-particle-irreducible (1PI); A4: two-loop 1PR; A5: two-loop (1PI). A diagram is reducible if cutting a single internal line disconnects the graph. Reducible topologies are less common in textbooks and thus test whether models rely on memorized patterns or perform genuine structural deduction.
    \item \textbf{Interaction sector (B1--B6):} B1: QED with photons and $e^\pm$ only; B2: QED with generic charged fermions; B3: electroweak with leptons and neutrinos; B4: electroweak including $W/Z$ gauge bosons; B5: electroweak including the Higgs boson; B6: strong (QCD) with quarks and gluons.
\end{itemize}

In total, we identify 102 distinct task types, produce over 200,000 candidate diagrams, and select a representative subset of 2,000 that preserves diversity across topology and interaction classes. Table~\ref{tab:topology_summary} reports the per-category counts. The benchmark is intentionally unbalanced across categories---certain higher-loop electroweak sectors naturally contain many more valid diagrams than simple QED trees---and we preserve this natural variation while ensuring that all topology and interaction classes remain represented.

\begin{table*}[t]
	\caption{Diagram counts by interaction sector and topology class. Topology: tree (A1), one-loop 1PR (A2), one-loop 1PI (A3), two-loop 1PR (A4), two-loop 1PI (A5).}
	\label{tab:topology_summary}
	\centering
	\footnotesize
	\setlength{\tabcolsep}{2pt}
	\renewcommand{\arraystretch}{1.15}
	\begin{tabular*}{\textwidth}{@{\extracolsep{\fill}} l l c c c c c}
		\toprule
		Interaction sector &
		External-leg content &
		A1 & A2 & A3 & A4 & A5 \\
		\midrule
		QED &
		Photons, $e^\pm$ only &
		15 & 86  & 17  & 150 & 119 \\
		QED &
		Photons + generic charged fermions &
		11 & 112 & 12  & 106 & 72  \\
		Electroweak &
		Leptons and neutrinos &
		45 & 125 & 51  & 50  & 120 \\
		Electroweak &
		Includes $W/Z$ &
		8  & 115 & 68  & 75  & 147 \\
		Electroweak &
		Includes Higgs &
		17 & 56  & 122 & 70  & 89  \\
		QCD &
		Quarks / gluons &
		7  & 38  & 74  & 51  & 28  \\
		\bottomrule
	\end{tabular*}
	\label{cat}
\end{table*}

\begin{figure*}[hbtp]
	\centering
	\includegraphics[width=\linewidth]{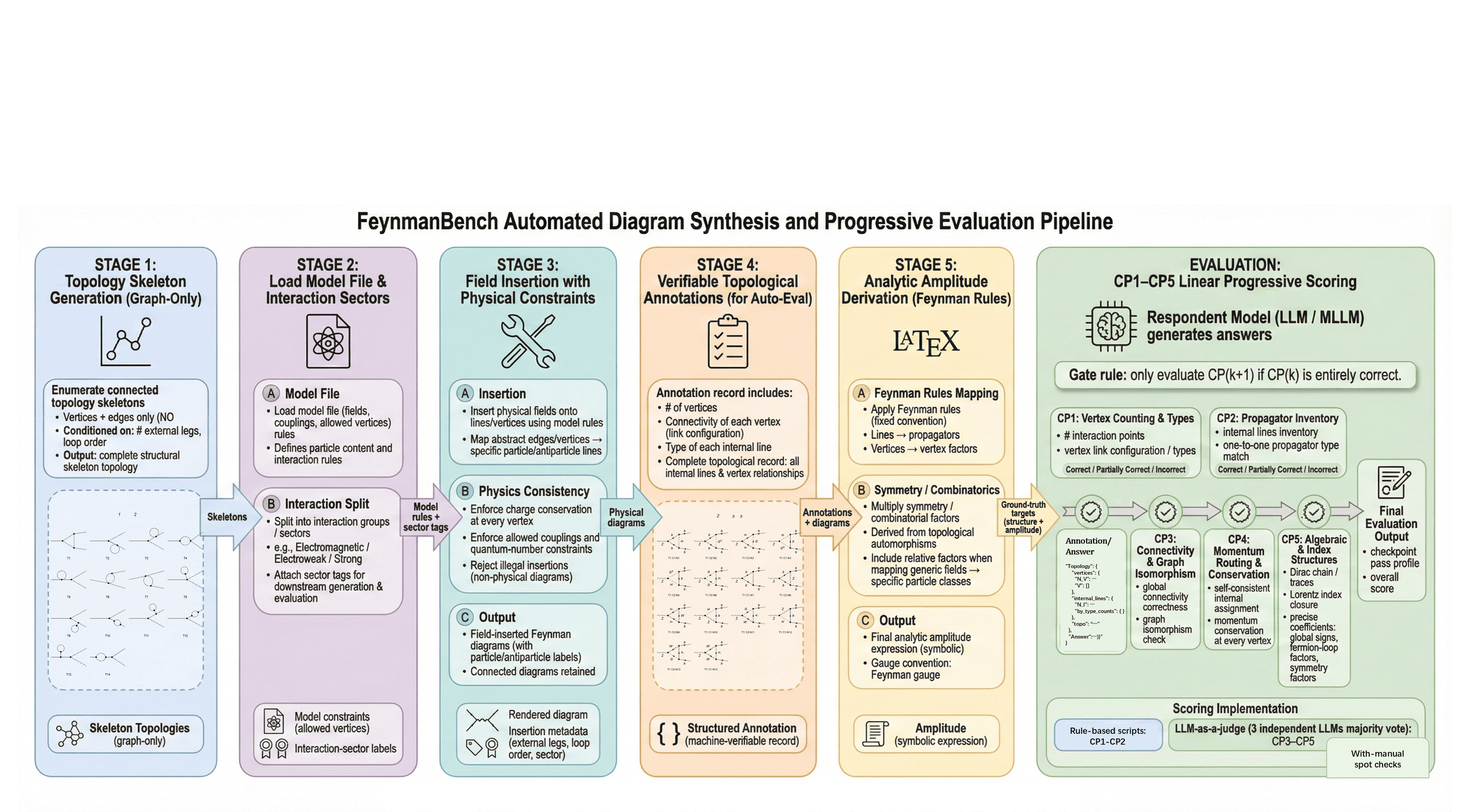}
	\caption{Workflow of FeynmanBench. Panels 1--4: diagram generation and annotation. Panel 5: amplitude derivation via Feynman rules. Panel 6: automated scoring against ground truth. The same pipeline that generates diagrams also produces their verifiable structural and symbolic ground truth, enabling scalable evaluation without manual answer authoring.}
	\label{fig:workflow}
\end{figure*}

\section{Experiments}

\begin{figure*}[h]
	\centering
	\begin{minipage}[t]{0.45\linewidth}
		\centering
		\vspace{0pt}
		\includegraphics[width=\linewidth]{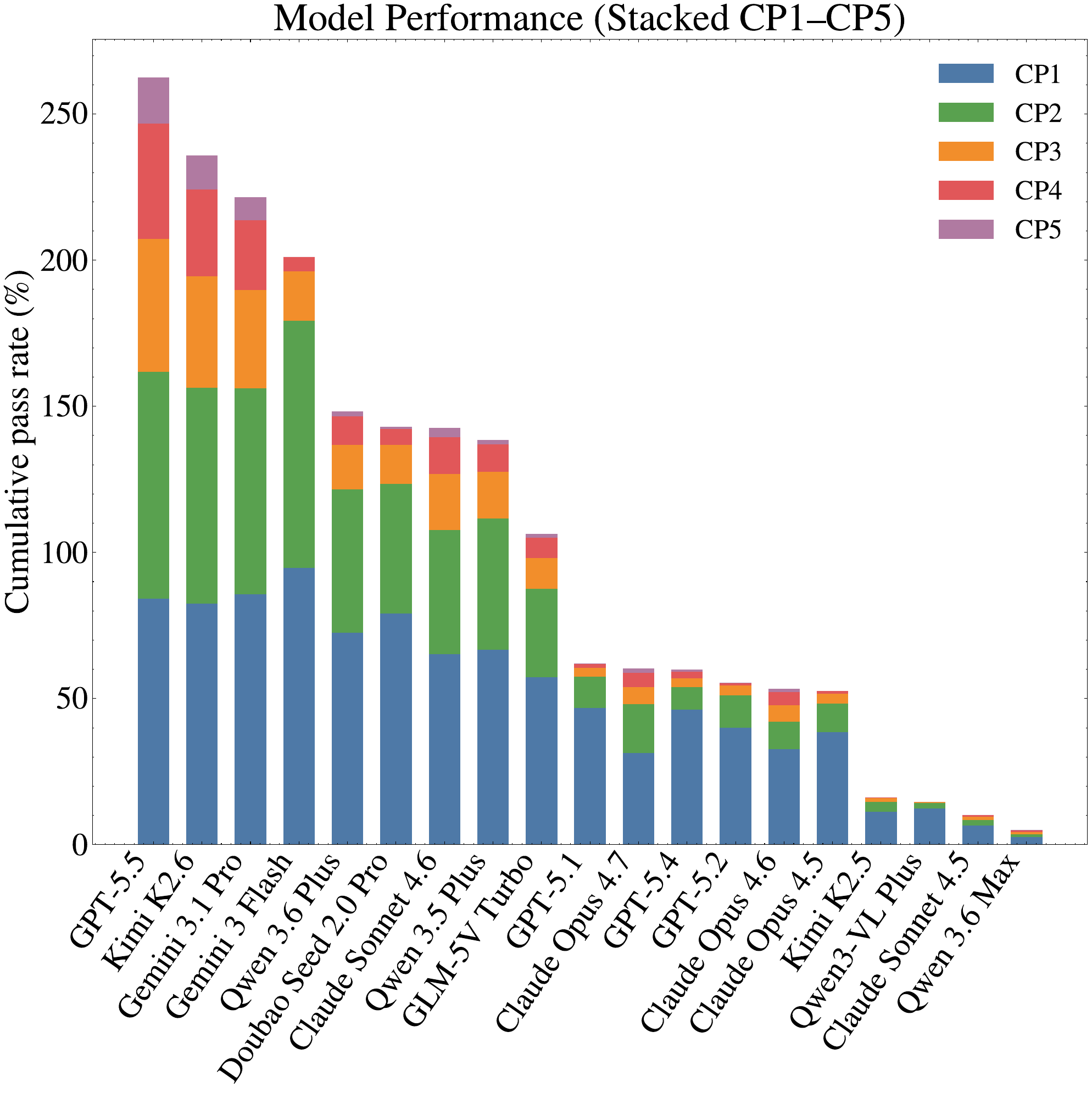}
	\end{minipage}\hfill
	\begin{minipage}[t]{0.45\linewidth}
		\centering
		\vspace{0pt}
		\includegraphics[width=\linewidth]{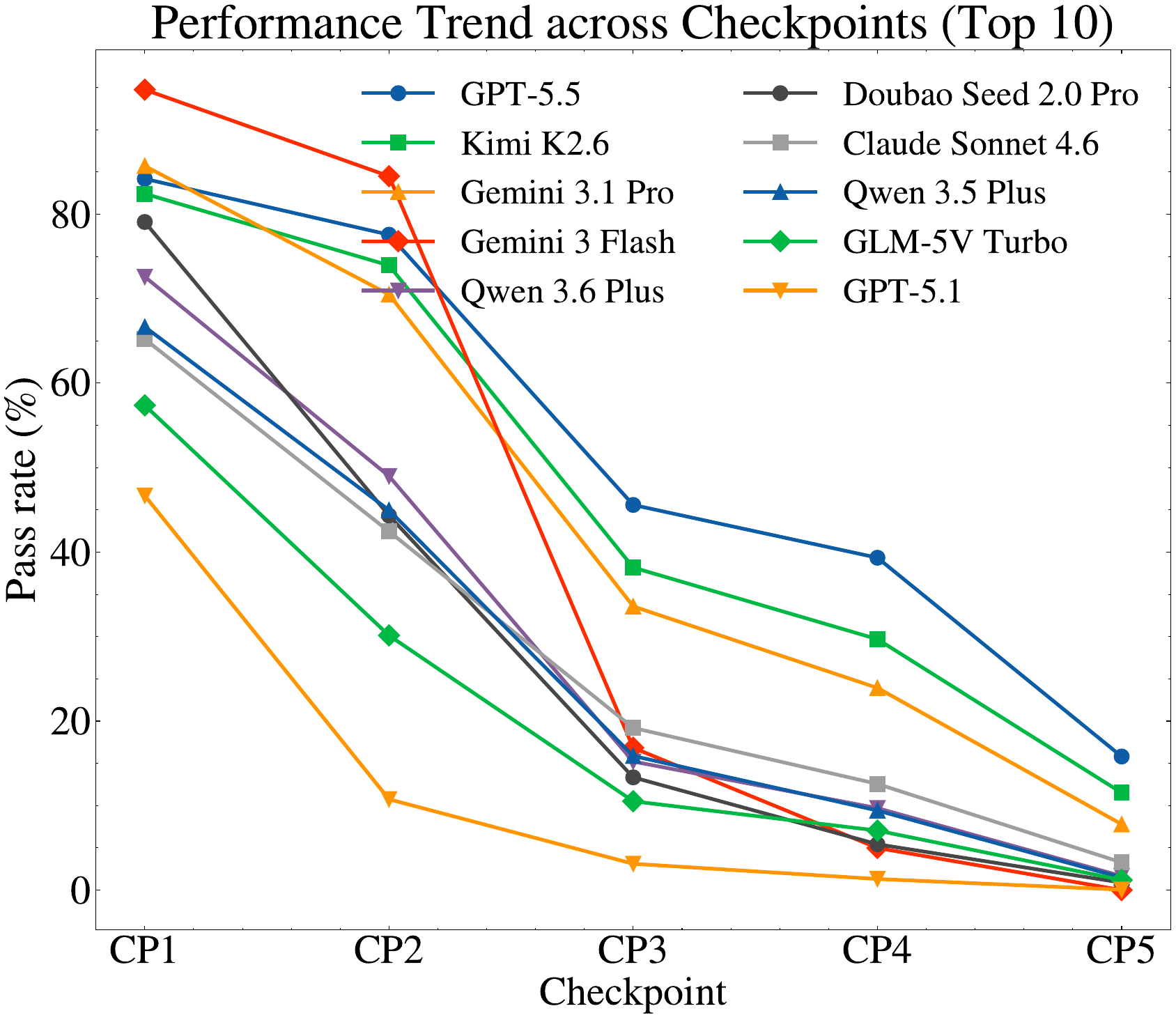}
	\end{minipage}
	\caption{Model performance across checkpoints.}
	\label{stockandtrend}
\end{figure*}

\subsection{Setup}

Figure~\ref{stockandtrend} and Table~\ref{tab:overall-cp} should be read primarily as evidence of a shared failure pattern rather than a leaderboard: across diverse model families and scales, local recognition remains substantially stronger than topology-sensitive reasoning.

\textbf{Models.} We evaluate 19 multimodal LLMs from five major providers, spanning the GPT, Gemini, Claude, Qwen, GLM, Kimi, and Doubao families. Due to API rate limits and cost constraints, we evaluate four models (GPT-5.5, Kimi~K2.6, Gemini~3.1~Pro, Gemini~3~Flash) on a representative subset ($N$ ranging from 165 to 742), stratified by topology and interaction classes to match the full benchmark distribution; the remaining 15 models are evaluated on the full benchmark ($N \approx 2{,}000$). All models are accessed via API with temperature set to zero. We perform five independent evaluation runs per instance and report the mean pass rate across runs. Although decoding is configured deterministically (temperature = 0), we retain repeated runs because closed-source API systems can exhibit mild nondeterminism due to backend routing and vision preprocessing pipelines; averaging across runs reduces such variance.

\begin{figure}[t]
	\centering
	\includegraphics[width=\columnwidth]{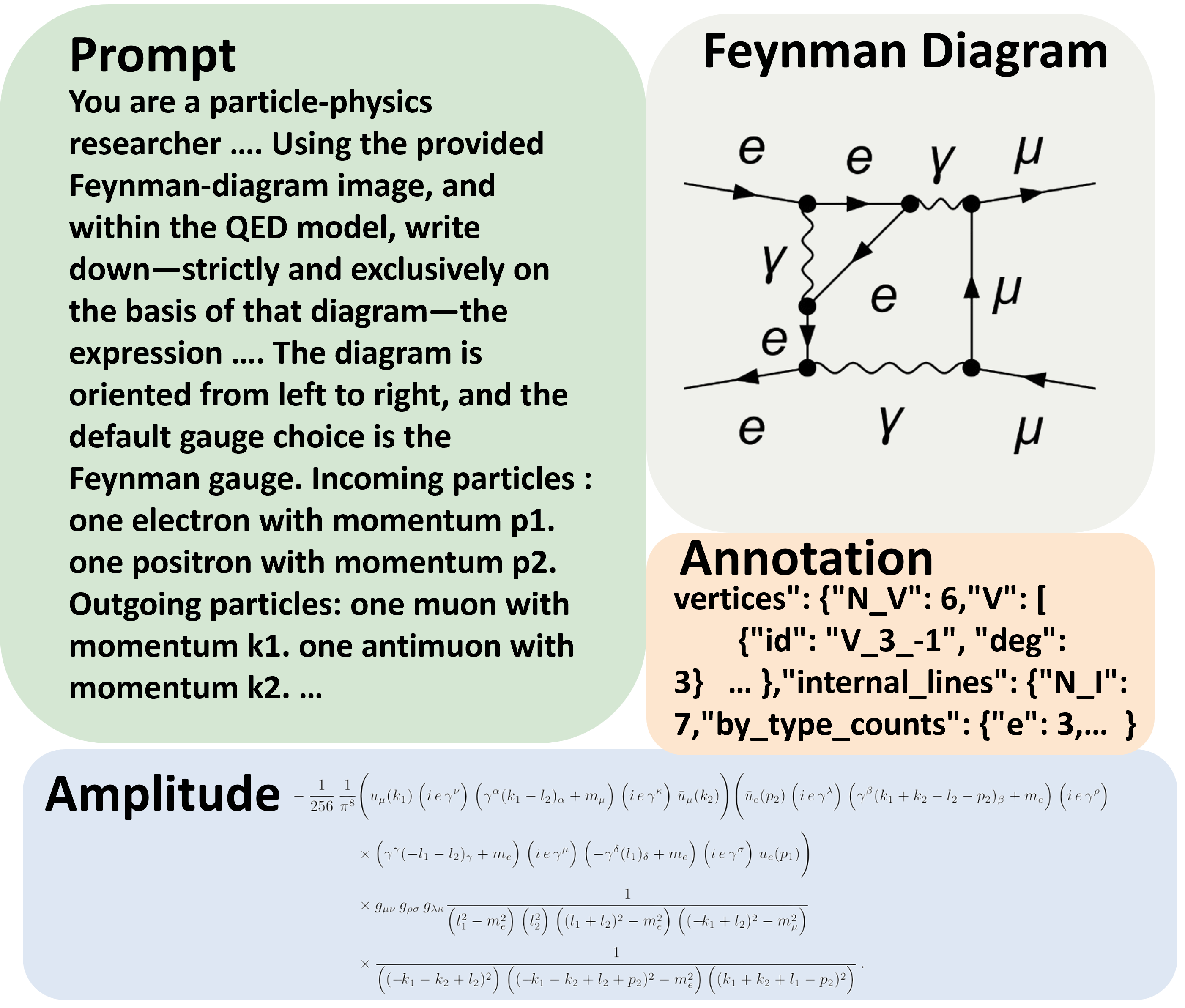}
	\caption{Prompt structure and task definitions with example expectations. The prompt provides minimal conventions (interaction type, external-leg identities, momentum orientation)---sufficient to resolve drawing ambiguities without supplying topological or algebraic answers.}
	\label{prompt}
\end{figure}

\textbf{Prompt.} As shown in Fig.~\ref{prompt}, we design the prompt to provide only the minimum conventions needed to make the task well-posed: interaction type, external-leg identities, momentum labels, and diagram orientation. These elements disambiguate drawing conventions (e.g.,~momentum-direction rules for antiparticles) without supplying any topological or algebraic answer. We intentionally provide only these metadata-level cues---the critical reasoning for CP3--CP5 must still be performed over the diagram structure itself.

\textbf{Evaluation protocol.} We evaluate the five checkpoints independently. For CP1 and CP2, we use automated rule-based scripts that verify correctness against ground-truth annotations, distinguishing three grades: correct, partially correct (right count but wrong type or linkage), and incorrect. For CP3--CP5, we adopt an LLM-as-a-judge framework with a fixed evaluation prompt, a panel of three independent models, and majority vote to determine the final score. To validate this protocol, we conduct manual spot-checks on 100 randomly sampled instances across CP3--CP5; we find 91\% agreement between human judgment and the majority-vote outcome, confirming the reliability of automated scoring.

\begin{table*}[t]
	\centering
	\small
	\setlength{\tabcolsep}{4pt}
	\caption{Overall checkpoint pass rates (\%). $\text{Score}=\frac{1}{5}\sum_j \text{CP}_j$. $\protect\dagger$: evaluated on a stratified subset. $\protect\dagger\protect\dagger$: not covered for this category.}
	\begin{tabular}{lrrrrrrr}
		\toprule
		Model & $N$ & CP1 & CP2 & CP3 & CP4 & CP5 & Score \\
		\midrule
		GPT-5.5$^\dagger$           & 272  & 84.2 & 77.6 & 45.6 & 39.3 & 15.8 & 52.5 \\
		Kimi K2.6$^\dagger$         & 165  & 82.4 & 73.9 & 38.2 & 29.7 & 11.5 & 47.2 \\
		Gemini 3.1 Pro$^\dagger$    & 539  & 85.7 & 70.5 & 33.6 & 23.9 & 7.8  & 44.3 \\
		Gemini 3 Flash$^\dagger$    & 742  & 94.7 & 84.5 & 16.8 & 5.0  & 0.0  & 40.2 \\
		Qwen 3.6 Plus       & 2051 & 72.6 & 49.0 & 15.2 & 9.7  & 1.7  & 29.6 \\
		Doubao Seed 2.0 Pro  & 2046 & 79.1 & 44.3 & 13.3 & 5.4  & 0.9  & 28.6 \\
		Claude Sonnet 4.6    & 2021 & 65.2 & 42.5 & 19.2 & 12.6 & 3.3  & 28.5 \\
		Qwen 3.5 Plus        & 2028 & 66.7 & 45.0 & 15.9 & 9.4  & 1.5  & 27.7 \\
		GLM-5V Turbo          & 1931 & 57.4 & 30.1 & 10.5 & 7.0  & 1.2  & 21.3 \\
		GPT-5.1               & 2056 & 46.7 & 10.7 & 3.1  & 1.3  & 0.0  & 12.4 \\
		Claude Opus 4.7       & 2052 & 31.4 & 16.6 & 5.9  & 4.7  & 1.5  & 12.0 \\
		GPT-5.4               & 2056 & 46.3 & 7.7  & 3.0  & 2.1  & 0.9  & 12.0 \\
		GPT-5.2               & 2056 & 39.9 & 11.2 & 3.3  & 0.9  & 0.0  & 11.1 \\
		Claude Opus 4.6       & 2046 & 32.6 & 9.4  & 5.7  & 4.5  & 1.0  & 10.7 \\
		Claude Opus 4.5       & 2052 & 38.6 & 9.6  & 3.5  & 0.9  & 0.0  & 10.5 \\
		Kimi K2.5             & 1871 & 11.2 & 3.4  & 1.4  & 0.2  & 0.0  & 3.2  \\
		Qwen3-VL Plus         & 2044 & 12.3 & 2.0  & 0.3  & 0.0  & 0.0  & 2.9  \\
		Claude Sonnet 4.5     & 2035 & 6.5  & 2.0  & 1.1  & 0.6  & 0.0  & 2.0  \\
		Qwen 3.6 Max          & 2053 & 2.5  & 1.0  & 0.8  & 0.5  & 0.2  & 1.0  \\
		\bottomrule
	\end{tabular}
	\label{tab:overall-cp}
\end{table*}

\subsection{Results Analysis}

We present the full evaluation results in Fig.~\ref{stockandtrend} and Table~\ref{tab:overall-cp}. Our analysis reveals three key findings.

\textbf{The CP3 cliff.} We observe a sharp and universal performance collapse from local recognition to topological reconstruction in every model tested. Gemini~3~Flash exemplifies the pattern: 94.7\% on CP1 and 84.5\% on CP2, dropping to 16.8\% on CP3 and 5.0\% on CP4. Doubao Seed~2.0~Pro scores 79.1\% on CP1 but only 13.3\% on CP3. We find this pattern across all model families and scales. This CP3 collapse is the central diagnostic of FeynmanBench: a model can achieve high CP1/CP2 rates by recognizing local symbols and line styles, but CP3 requires constructing a globally coherent latent graph and verifying that distant regions of the image are mutually compatible. The collapse therefore indicates a failure of structural composition rather than a mere loss of low-level visual accuracy. Because four frontier models are evaluated on stratified subsets rather than the full benchmark, we focus our conclusions on performance trends across checkpoints rather than fine-grained leaderboard comparisons.

\begin{figure}[htbp]
	\centering
	\includegraphics[width=0.95\columnwidth]{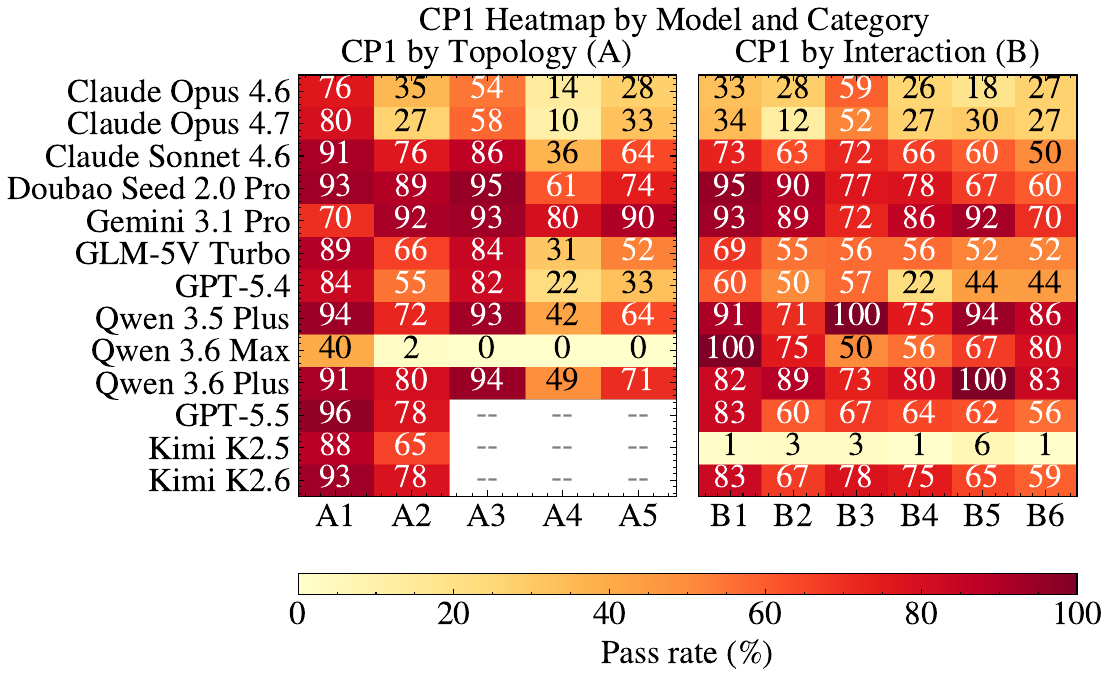}\par
	\vspace{0.3em}
	\includegraphics[width=0.95\columnwidth]{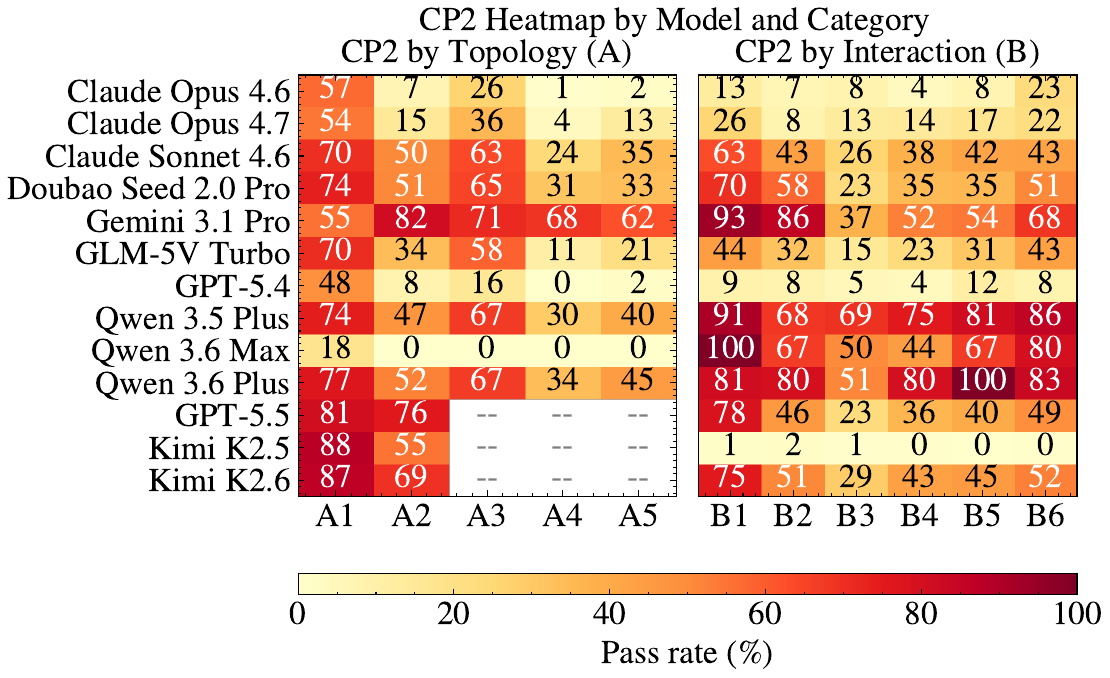}\par
	\vspace{0.3em}
	\includegraphics[width=0.95\columnwidth]{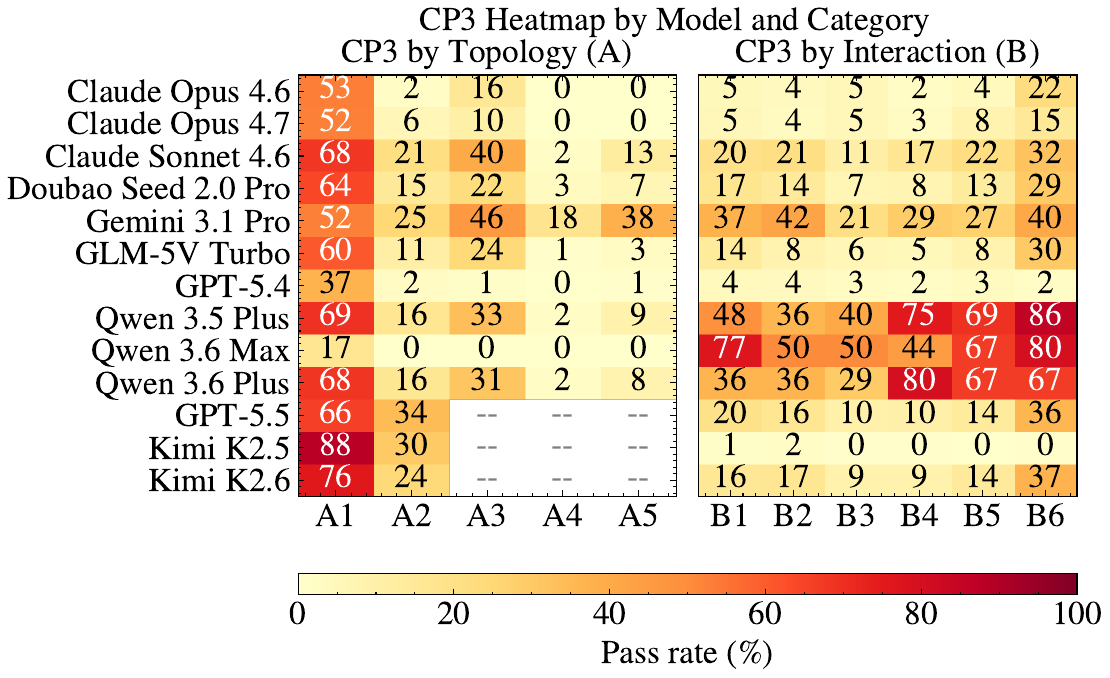}
	\caption{Heatmaps by model and category for CP1--CP3.}
	\label{fig:cp1_cp3_heatmaps_vertical}
\end{figure}

\textbf{Topology complexity and memorization.} We stratify CP4 pass rates by topology class (Table~\ref{tab:cp4_by_A}) and find sharp degradation with increasing complexity: even top models achieve 40--57\% on tree-level diagrams (A1) but collapse to near zero on two-loop structures (A4, A5). Critically, we observe that reducible graphs (A2) are consistently harder than irreducible ones (A3)---Claude~Sonnet~4.6 scores 16.2\% vs.\ 28.2\%; Qwen~3.6~Plus scores 11.0\% vs.\ 21.4\%. This pattern is counterintuitive from a graph-theoretic perspective, since 1PR graphs are often decomposable into simpler substructures. Their greater difficulty for MLLMs suggests that rarity in training data outweighs compositional simplicity---models rely on memorized patterns from textbooks rather than transferable structural reasoning. We also stratify by interaction type: strong interactions (B6) yield the highest CP4 rates (23.9\% for Claude~Sonnet~4.6, 24.2\% for Qwen~3.6~Plus), while electroweak processes with leptons and neutrinos (B3) are consistently the hardest across all models, reflecting the added complexity of flavor mixing and chiral structures. The heatmaps in Fig.~\ref{fig:cp1_cp3_heatmaps_vertical} provide a finer-grained view of CP1--CP3 performance across individual topology and interaction categories, confirming that model strengths in local vertex recognition largely disappear when spatial complexity increases.

\textbf{CP5: near-zero algebraic reasoning.} We find that no model achieves meaningful CP5 performance---the best (GPT-5.5) reaches only 15.8\%, and all full-benchmark models score below 2\%. CP5 requires the complete derivation of symmetry factors, Dirac matrix orderings, and overall normalization. Examining the rare CP5 successes, we note that they correspond almost exclusively to diagrams found in published textbooks, further supporting our memorization hypothesis. CP5 is especially demanding because errors compound from earlier stages: an incorrect propagator type from CP2, a missing vertex connection from CP3, or an inconsistent momentum route from CP4 each independently invalidate the final amplitude, even if the symbolic form appears superficially plausible. The near-zero CP5 rates therefore reflect not only symbolic weakness but also the absence of robust graph-level internal representations upstream.

\subsection{Error Analysis}

\begin{figure*}[h]
	\centering
	\includegraphics[width=0.95\textwidth]{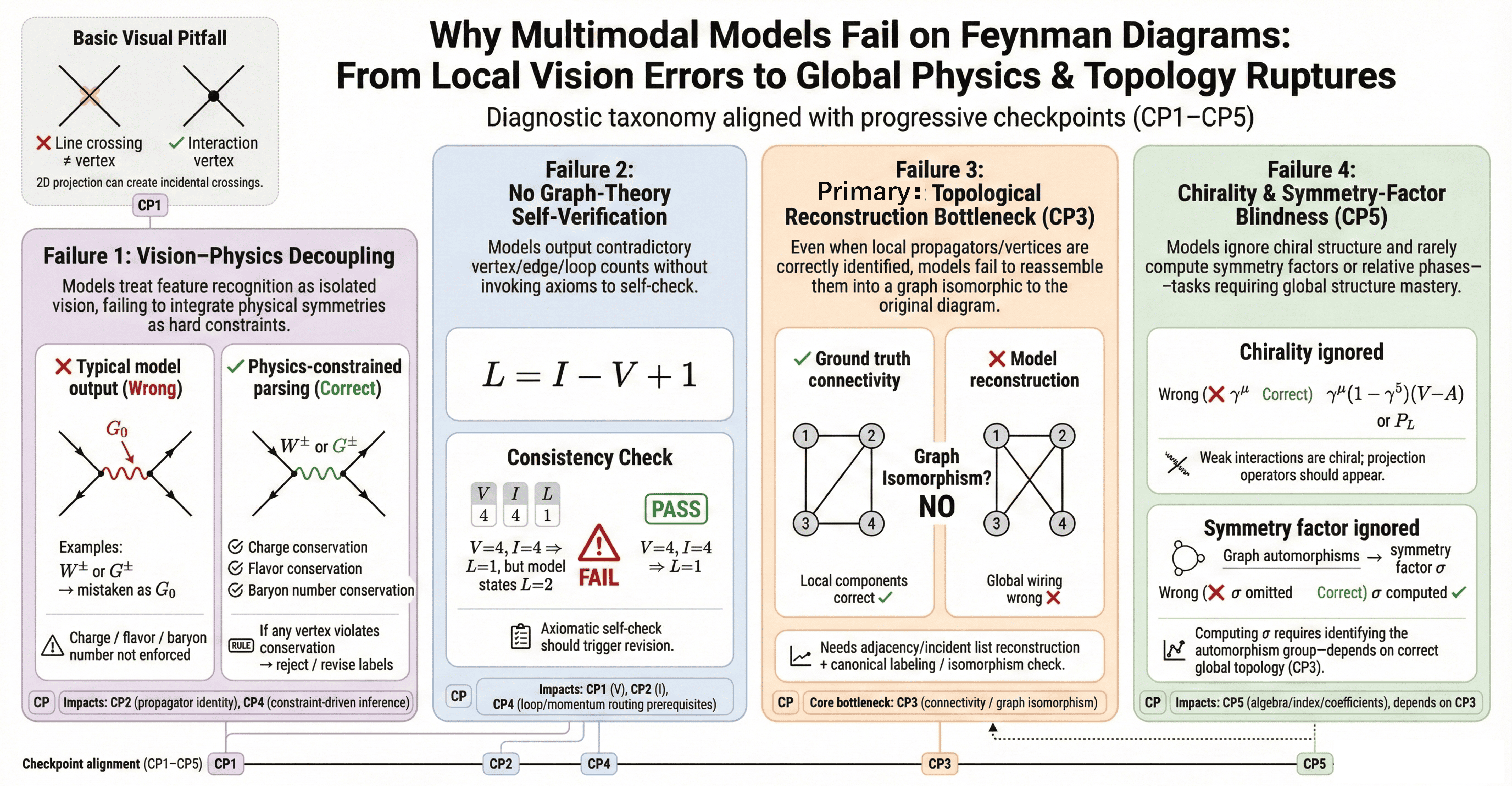}
	\caption{Representative examples of the four dominant failure modes. Each case corresponds to a pattern repeatedly observed across models and interaction sectors, rather than an isolated anomalous output.}
	\label{classerror}
\end{figure*}

Through qualitative analysis of model outputs, we identify four recurring failure modes. These errors are not independent: in many cases, an early visual parsing mistake triggers a cascade---a spurious vertex changes the inferred propagator set, which in turn breaks graph connectivity, invalidates momentum conservation, and makes the final amplitude unsalvageable. This cascading pattern helps explain why CP5 remains near zero even when CP1 and CP2 scores are moderately strong. Figure~\ref{classerror} illustrates representative examples:

\begin{itemize}
    \item \textbf{Visual parsing failures.} We find that models frequently mistake incidental line crossings in 2D projections for interaction vertices, inflating vertex counts and corrupting downstream topological reconstruction.
    \item \textbf{Vision-logic decoupling.} When inferring internal-line identities, models fail to apply conservation laws (charge, flavor, baryon number). We consistently observe that charged bosons ($W^\pm$, $G^\pm$) are conflated with neutral counterparts ($G^0$, $Z$), even though the external-leg charge configuration determines the internal identity uniquely.
    \item \textbf{Graph-theoretic inconsistency.} The topological identity $L = I - V + 1$ (relating loops $L$, internal lines $I$, and vertices $V$) is routinely violated in model outputs. We note that models report $V$, $I$, and $L$ values that cannot simultaneously satisfy this basic graph-theoretic constraint, indicating a fundamental lack of self-consistency checking.
    \item \textbf{Ignoring symmetry and chirality.} We observe that models make almost no attempt to compute combinatorial symmetry factors or to invoke chiral projection operators ($V-A$ structures), even when explicitly prompted. We believe determining a symmetry factor requires identifying the automorphism group of the graph---a task demanding precise global structural mastery that current architectures lack. This failure mode directly explains why CP5 remains near zero: even a correctly reconstructed topology cannot yield a correct amplitude without the proper symmetry factor.
\end{itemize}

\begin{table}[t]
	\centering
	\small
	\setlength{\tabcolsep}{3pt}
	\renewcommand{\arraystretch}{1.12}
	\caption{CP4 pass rates (\%) by topology class (A1--A5), isolating the effect of graph complexity. Without this stratification, aggregate scores would obscure the fact that most models fail almost completely once reasoning requires multi-loop global consistency. $\protect\dagger$: subset. ``--'' denotes category not covered.}
	\label{tab:cp4_by_A}
	\resizebox{\columnwidth}{!}{\begin{tabular}{lrrrrrr}
		\toprule
		Model & A1 & A2 & A3 & A4 & A5 & Overall \\
		\midrule
		Kimi K2.5$^\dagger$       & 75.0 & 25.0 & --   & --   & --   & 52.3 \\
		GPT-5.5$^\dagger$         & 57.6 & 28.9 & --   & --   & --   & 39.3 \\
		Kimi K2.6$^\dagger$       & 53.3 & 20.8 & --   & --   & --   & 29.7 \\
		Gemini 3.1 Pro$^\dagger$  & 39.8 & 18.0 & 32.2 & 8.2  & 27.6 & 23.9 \\
		Claude Sonnet 4.6  & 56.9 & 16.2 & 28.2 & 0.2  & 3.7  & 12.6 \\
		Qwen 3.6 Plus      & 48.5 & 11.0 & 21.4 & 0.6  & 3.3  & 9.7  \\
		Qwen 3.5 Plus      & 47.5 & 10.5 & 21.3 & 0.2  & 3.4  & 9.4  \\
		GLM-5V Turbo       & 42.9 & 8.5  & 15.0 & 0.2  & 0.7  & 7.0  \\
		Doubao Seed 2.0 Pro & 48.5 & 5.8  & 8.7  & 0.0  & 0.3  & 5.4  \\
		Claude Opus 4.7     & 45.6 & 5.2  & 6.7  & 0.0  & 0.0  & 4.7  \\
		Claude Opus 4.6     & 44.0 & 1.9  & 11.7 & 0.0  & 0.5  & 4.5  \\
		GPT-5.4             & 26.2 & 2.1  & 1.0  & 0.0  & 0.3  & 2.1  \\
		\bottomrule
	\end{tabular}}
	\vspace{2pt}
	\parbox{\columnwidth}{\footnotesize\textit{Note:} Each rate is $\#\text{(CP4 correct)} / \#\text{(total items in category)}$.}
\end{table}

\section*{Limitations}

We acknowledge several limitations of the current work. First, FeynmanBench covers Standard Model perturbative diagrams up to two-loop order under Feynman gauge; it does not include gravitational interactions, non-perturbative regimes, or higher-loop contributions. Second, we evaluated four frontier models on stratified subsets ($N = 165$--$742$) rather than the full benchmark; while the subset preserves category-level distributions, these smaller samples reduce statistical resolution for subgroup analyses. We also note that our LLM-as-a-judge protocol for CP3--CP5 was spot-checked on 100 instances (91\% agreement with majority vote) but has not been independently verified across every instance, and scoring sensitivity to prompt variations remains unexamined. Third, all diagrams are machine-generated under standardized rendering; model behavior on hand-drawn or photographed diagrams with visual noise may differ. Finally, our benchmark evaluates single-pass diagram comprehension under fixed textual conventions, and our model coverage is limited to proprietary API-accessible systems without open-weight models or human expert baselines for calibration.

\section{Conclusion}

We introduced FeynmanBench, a benchmark of over 2,000 Feynman-diagram tasks with automated generation and verification, designed to evaluate multimodal LLMs on diagrammatic reasoning that requires global structural understanding. Evaluating 19 models reveals a consistent failure mode: local visual recognition is relatively strong, but performance collapses at topological reconstruction (CP3) and is virtually absent at full algebraic derivation (CP5). The inversion of reducible vs.~irreducible difficulty further suggests reliance on memorization over structural deduction. These results indicate that the primary bottleneck for formal scientific diagrams is not visual perception but the maintenance of globally consistent representations under strict constraints. We believe this gap between local parsing and global reasoning will persist for any notation system whose semantics are defined relationally rather than compositionally. We will release our code, pipeline, and benchmark upon acceptance. Future work includes extending the benchmark to higher-loop orders and effective field theories, as well as evaluating the inverse generative task of synthesizing correct Feynman diagrams from textual physics descriptions.


\bibliographystyle{ACM-Reference-Format}

\end{document}